\documentclass[runningheads]{llncs}

\usepackage{eccv}

\usepackage[T1]{fontenc}

\usepackage[pagebackref,breaklinks,colorlinks]{hyperref}

\usepackage{times}
\usepackage{enumitem}

\usepackage[utf8]{inputenc}
\usepackage[english]{babel}
\usepackage{amsmath}
\usepackage{amsfonts}
\usepackage{amssymb}
\usepackage{graphicx}
\usepackage{bm} 

\usepackage[dvipsnames]{xcolor}

\usepackage{caption}

\usepackage{wrapfig} 

\usepackage{float}

\usepackage{pifont}
\newcommand{\xmark}{\color{purple} \ding{55}}
\newcommand{\xmarkbw}{\ding{55}}
\newcommand{\cmark}{\color{olive} \ding{51}}

\usepackage{tablefootnote}

\title{YCB-Ev SD: Synthetic event-vision dataset for 6DoF object pose estimation}
\author{Pavel Rojtberg\orcidID{0000-0001-9736-6866} \and Julius Kühn\orcidID{0000-0003-2458-0030}}

\makeatletter

\hypersetup{
    pdftitle={\@title}
}

\begin{document}
	    \institute{Fraunhofer IGD, Darmstadt, Germany \email\{pavel.rojtberg, julius.kuehn\}@igd.fraunhofer.de}
	\maketitle
\begin{abstract}
 	We introduce YCB-Ev SD, a synthetic dataset of event-camera data at standard definition (SD) resolution for 6DoF object pose estimation. While synthetic data has become fundamental in frame-based computer vision, event-based vision lacks comparable comprehensive resources. Addressing this gap, we present 50,000 event sequences of 34 ms duration each, synthesized from Physically Based Rendering (PBR) scenes of YCB-Video objects following the Benchmark for 6D Object Pose (BOP) methodology. Our generation framework employs simulated linear camera motion to ensure complete scene coverage, including background activity.
 
 	Through systematic evaluation of event representations for CNN-based inference, we demonstrate that time-surfaces with linear decay and dual-channel polarity encoding achieve superior pose estimation performance, outperforming exponential decay and single-channel alternatives by significant margins. Our analysis reveals that polarity information contributes most substantially to performance gains, while linear temporal encoding preserves critical motion information more effectively than exponential decay. 
 	The dataset is provided in a structured format with both raw event streams and precomputed optimal representations to facilitate immediate research use and reproducible benchmarking.
	
    The dataset is publicly available at \url{https://huggingface.co/datasets/paroj/ycbev_sd}.
    
\end{abstract}

	\begin{figure}
		\centering
		\captionsetup{type=figure}
		\includegraphics[width=0.32\linewidth]{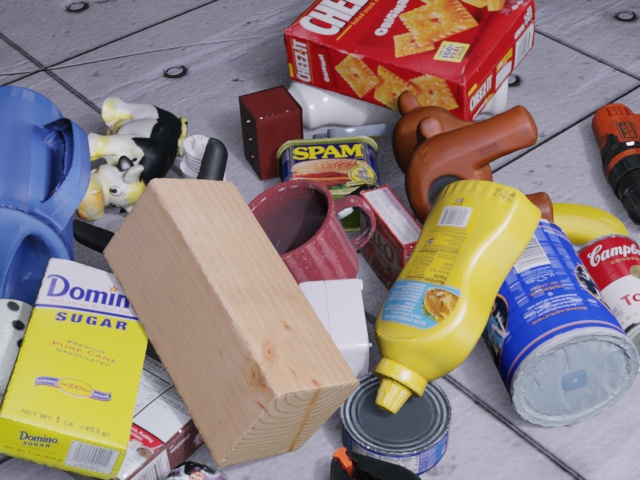}
		\includegraphics[width=0.32\linewidth]{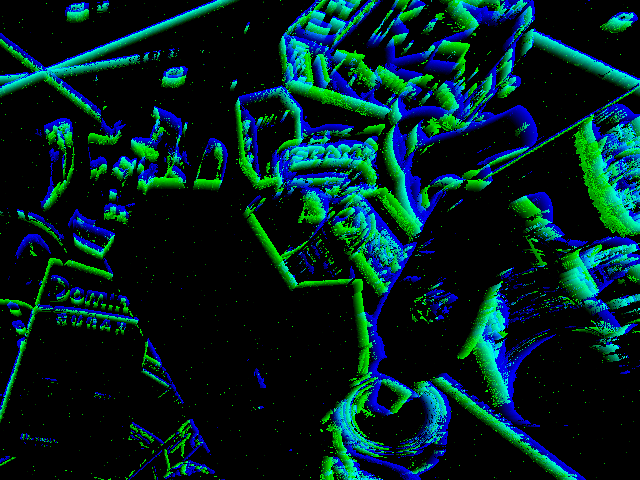}
		\includegraphics[width=0.32\linewidth]{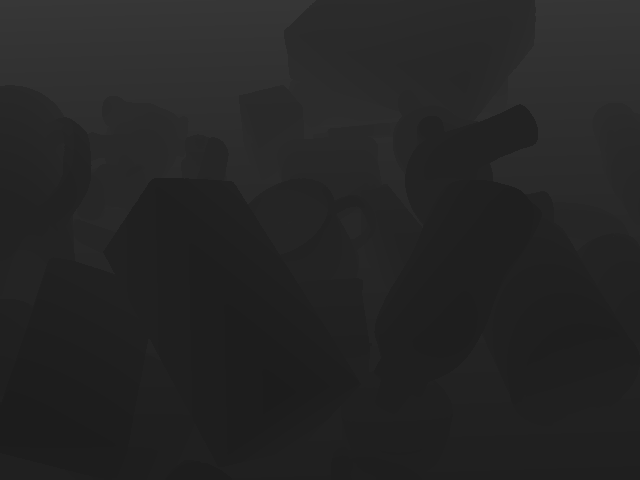}
		\captionof{figure}{
        Sythethic RGB renderings (left), matching event histogram (middle), and depth image (right) from the our dataset.    
        Throughout this paper, the polarity of the events is represented as blue (falling) and green (rising).}
		\label{fig:teaser}
	\end{figure}%

    \section{Introduction and related work}
    Real-time object detection and 6D pose estimation are critical capabilities for applications in augmented reality, virtual reality, and robotics. The field's progress is benchmarked by the BOP Challenge \cite{nguyen2025bop}, which annually ranks algorithms and publishes a leaderboard. This challenge utilizes a variety of datasets that differ in captured modality (e.g., RGB, depth) and object types (e.g., household, industrial). Since 2020, the challenge has included 50,000 synthetic, physically based rendering (PBR) images, which have proven highly beneficial. These images provide precise annotations and a wide variety of lighting conditions and scene arrangements, often boosting performance beyond using real data alone.
    
    Among the datasets used in the BOP challenge, the YCB-Video (YCB-V) dataset \cite{xiang2018posecnn} is particularly notable. It not only provides 3D models for generating synthetic renderings but also allows researchers to obtain the corresponding physical objects from the YCB organizers \cite{calli2017yale}.
    
    Event-based, or neuromorphic, cameras offer a novel sensing modality with advantages over traditional frame-based cameras, including high temporal resolution, high dynamic range, and low power consumption. Instead of capturing full frames at a fixed rate, they output a sparse, asynchronous stream of "events." Each event is triggered at a pixel when the intensity change exceeds a threshold, encoding the location, timestamp, and polarity (sign) of the change (see Figure \ref{fig:teaser}).

 \begin{table*}
 	\centering
 	\begin{tabular}{|c|c|c|c|c|c|}
 		\hline
 		Dataset & object classes & frames & high-res $^1$ & real data & synthetic data \\
 		\hline
 		\hline
 		YCB-Ev \cite{rojtberg2024ycb} & 21 & 13,851 & \cmark & \cmark & \xmark \\
 		\hline
 		E-POSE \cite{hay2025pose} & 13 & 333,357 & \xmark & \cmark & \xmark \\
 		\hline
 		YCB-Ev SD (ours) & 21 & 50,000 & \xmark & \xmark & \cmark \\
 		\hline
 	\end{tabular}
 	\caption{Overview of related YCB-based event datasets. Our dataset is distinguished as the only one providing synthetic event data.}
 	\footnotesize{$^1$ High-resolution is defined as $\geq$ HD (1280x720px)}\\
 	\label{tbl:ycbdatasets}
 \end{table*}
 
    \autoref{tbl:ycbdatasets} provides an overview of existing YCB-based event datasets. Current datasets primarily focus on real event data, mirroring the initial approach taken with RGB-D data. For synthetic event data dedicated to object pose estimation, to the best of our knowledge, only one satellite-centric dataset exists \cite{yishi2025cross}. However, this dataset uses simple OpenGL rendering instead of more realistic PBR, and its scenes contain no background and only one satellite per image.

	In this work, we align with the BOP challenge's methodology for RGB data by providing 50,000 synthetic event sequences generated from PBR scenes. Existing real event datasets offer either a low resolution of 346x260px \cite{hay2025pose} or a high HD resolution of 1280x720 \cite{rojtberg2024ycb}. Recent studies \cite{gehrig2022high} suggest that high-resolution event cameras can exhibit lower task performance in low-light conditions or during fast motion due to increased temporal noise. Specifically, \cite{gehrig2022high} demonstrated that standard definition (SD) cameras (640x360) outperformed HD cameras in camera pose tracking. Therefore, we also adopt SD resolution for our dataset. This choice reduces storage requirements and facilitates direct comparison with the BOP PBR images, which also use SD resolution.

	Additionally, we investigate event data representations suitable for processing with Convolutional Neural Networks (CNNs). While less computationally efficient than Spiking Neural Networks (SNNs), CNNs currently offer superior inference speeds \cite{mechler2023transferring}.

    Based on the above, our key contributions are;
    \begin{enumerate}[nolistsep]
 	  \item A large-scale synthetic event dataset for object pose estimation, and
 	  \item a comparative analysis of event representations for CNN-based inference.
    \end{enumerate}
 
    This paper is structured as follows: Section \ref{sec:datagen} details our synthetic data generation pipeline.
    Section \ref{sec:eventrep} discusses and evaluates different event data representations for CNN processing, while Section \ref{sec:datafmt} explains the structure and storage of the captured data.
    We conclude with Section \ref{sec:conclusion}, which summarizes our results, discusses the limitations, and outlines potential future work.

    \section{Synthetic event-data generation}
    \label{sec:datagen}
    
    Our synthetic data generation follows the BOP20 methodology \cite{hodavn2020bop}, producing a dataset of 50,000 distinct camera views using BlenderProc \cite{denninger2020blenderproc}.
	The views are organized into sequences of 25, with each sequence capturing the same object arrangement from a set of random camera poses.
    
	To synthesize event data corresponding to each camera pose, we introduce controlled camera motion, ensuring event generation across both foreground objects and background elements. For each target pose, we define a linear motion trajectory originating from a randomly sampled point on a sphere surrounding the target position and extending linearly toward it.
    
	We simulate this motion in BlenderProc over a 34 ms duration, rendering images at 1000 fps with a constant camera velocity of 1 cm/ms. The resulting high-frame-rate image sequences are subsequently converted into event streams using the IEBCS event camera simulator \cite{joubert2021event}.
	
	The event sequences generated through this process are sufficiently long to be accumulated into histograms at 30 fps (33 ms per frame), enabling direct comparison with conventional RGB video streams. This accumulation strategy facilitates fair benchmarking against standard frame-based approaches.

	The complete generation pipeline required rendering approximately 1.7 million high-quality PBR frames, equivalent to 28 minutes and 20 seconds of video at the simulated 1000 fps rate. This computationally intensive process required approximately one week utilizing six modern GPUs.

    \section{Comparing event representations for CNN inference}
    \label{sec:eventrep}
    
	Although Spiking Neural Networks (SNNs) offer computational advantages for processing event data, recent studies indicate that conventional CNN architectures still achieve superior frame rates \cite{mechler2023transferring}. Consequently, we evaluate various event representations \cite{gallego2020event} for their compatibility with CNN-based processing.
 
 	 \begin{figure}
 	\includegraphics[width=0.49\textwidth]{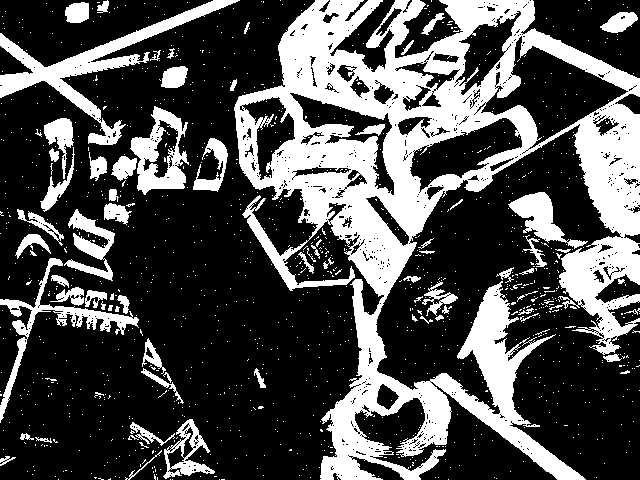}
 	\includegraphics[width=0.49\textwidth]{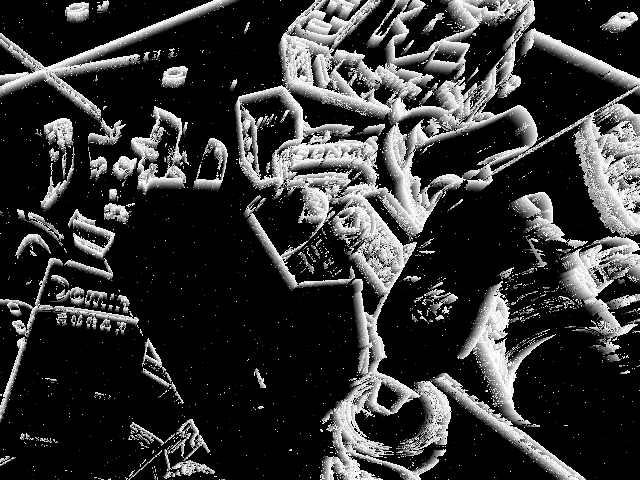}
 	\caption{Common event representations: (left) event histogram and (right) time-surface}
 	\label{fig:typical_evrep}
 	\end{figure}
 
 	We begin with event histograms and time-surfaces, as these map naturally to 2D grid structures familiar from image processing, making them widely adopted in literature \cite{jiao2021comparing} (see Figure \ref{fig:typical_evrep}).
   
    \begin{figure}
	\includegraphics[width=0.49\textwidth]{figure/000048.png}
	\includegraphics[width=0.49\textwidth]{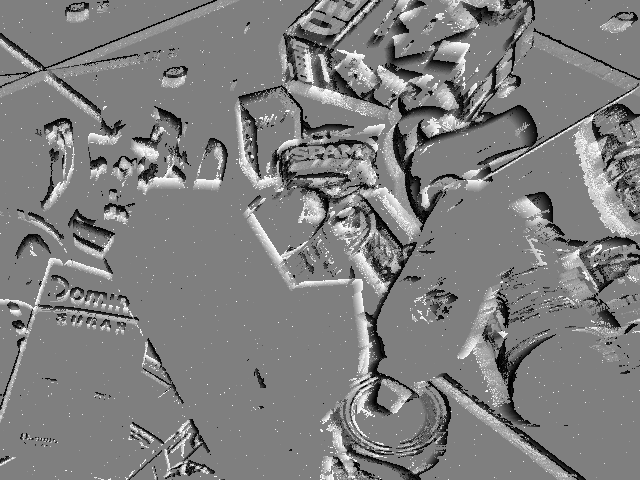}
	\caption{Time-surface representations with polarity encoding: (left) 2 channels in blue-green (right) 1 channel (black = off, white = on)}
	\label{fig:polarity_evrep}
	\end{figure}

	Time-surfaces typically employ an exponential decay function. We formulate this as:

	\begin{equation}
		I(\mathbf{x}, t) = 255 \cdot exp(-4 \cdot \dfrac{t_{max} - t}{t_{max}})
	\end{equation}
	
	where $t_{\text{max}} = 33$ ms represents the maximum time delta in the time-surface. The decay rate is empirically multiplied by 4 to better utilize the 8-bit value range when stored as image data.
	
	Since computing exponential decay on embedded hardware can be computationally expensive, and considering that neural networks can potentially learn this transformation, we also investigate a linear decay variant:

	\begin{equation}
	I(\mathbf{x}, t) = 255 \cdot \dfrac{t_{max} - t}{t_{max}}
	\end{equation}

	Event polarity can be incorporated into the representation through multiple encoding schemes: using separate color channels for each polarity, or a single channel where black represents off-events, white represents on-events, and gray indicates no activity (see Figure \ref{fig:polarity_evrep}). For time-surfaces, the single-channel approach halves the range available for representing temporal information, as both polarity histories must share one channel.

   	Given that event data is typically processed at high equivalent frame rates on resource-constrained embedded devices, we selected YOLOX6D-s \cite{maji2024yolo} for pose estimation --- a model capable of achieving 30 ms inference times on an NVIDIA Jetson Orin Nano.All experiments were conducted on a desktop GPU for development efficiency.

	We partitioned our dataset into 40,000 training, 5,000 validation, and 5,000 testing images. Each model was trained for 300 epochs from COCO-pretrained weights, retaining the checkpoint with lowest validation error.

	\begin{table*}
	\centering
	\begin{tabular}{|l|c|c|c|}
		\hline
		Representation & Event Polarity & Time Decay & ADD(-S) \\
		\hline
   		\hline
		Event Histogram & \xmarkbw & \xmarkbw & 8.4 \\
		\hline
		Event Histogram & 2 ch & \xmarkbw &  17.0 \\
		\hline
		Time Surface & \xmarkbw  & exp &  20.7 \\
		\hline
		Time Surface & \xmarkbw & linear &  23.7 \\
		\hline
		Time Surface & 2 ch & linear &  \textbf{30.4} \\
		\hline
		Time Surface & 1 ch & linear &  23.5 \\
		\hline
	\end{tabular}
	\caption{Performance comparison of different event representations on our synthetic test set using the ADD(-S)0.5d metric}
	\label{tbl:evaluation}
	\end{table*}

	Table \ref{tbl:evaluation} summarizes our evaluation results. The combination of polarity and temporal information substantially improves network performance, with polarity encoding demonstrating the strongest individual impact --- increasing accuracy by 8.6 percentage points for histograms and 6.7 percentage points for time surfaces.

	Temporal encoding resolution proves critical: exponential decay compresses recent values while extending older ones, and single-channel encoding halves the available dynamic range. Both approaches underperform compared to linear mapping with dual channels, which achieves optimal performance by preserving full temporal resolution and polarity information.

    \section{Dataset structure and usage}
	\label{sec:datafmt}
	This section details the data format and organization of our dataset. We adopt the BOP dataset structure for consistency, with extensions to accommodate event-based data modality.
	
	It is important to note that while we only provide ground truth 6D pose data, additional annotations such as 2D and 3D bounding boxes and per-pixel segmentation can be easily generated from the available 3D meshes by performing rasterization using the provided poses.
	
	Each captured sequence is stored in a subfolder that contains the sequence data in the format specified by the BOP dataset. The subfolder contains the following contents:
	\begin{itemize}[noitemsep]
		\item \texttt{rgb}: RGB color images in JPEG format
		\item \texttt{depth}: Per-pixel depth maps (millimeters)
		\item \texttt{ev\_raw}: Raw event data corresponding to the camera trajectory leading to each RGB-D pose (see Section \ref{sec:datagen})
		\item \texttt{ev\_histogram}: 2D time surfaces with linear decay (PNG format).
		\item \texttt{scene\_gt.json}: Ground truth poses for RGB frames
	\end{itemize}

	Event streams are stored in the \texttt{ev\_raw} subfolder using the compact, cross-platform format from \cite{rojtberg2024ycb}.
	
	Additionally, we provide precomputed event histograms in the \texttt{ev\_histogram} subfolder, corresponding to the optimal representation identified in Section \ref{sec:eventrep} for CNN-based processing.

	This histogram representation serves as a versatile base format that can be easily transformed into other discussed representations --- either by applying exponential decay mapping or by discarding polarity information as needed for specific applications.

    \section{Conclusion and future work}
    \label{sec:conclusion}
    
	This work introduces a large-scale synthetic dataset for 6D object pose estimation using event cameras. Our generation methodology employs simulated linear camera motion to ensure comprehensive scene coverage, including background activity, creating a rich and challenging benchmark for event-based vision.
    
    Through systematic evaluation of event representations for CNN-based processing, we demonstrate that time-surfaces with linear decay and dual-channel polarity encoding achieve superior pose estimation performance, outperforming exponential decay and single-channel alternatives by significant margins. Our analysis reveals that polarity information provides the strongest individual performance boost, while linear temporal encoding preserves motion information more effectively than exponential decay.
    
    To support immediate adoption and further research, we provide the dataset in a structured format containing both raw event streams and precomputed optimal histograms. This versatile foundation enables researchers to experiment with alternative event encodings without the computational overhead of reprocessing raw data.
    
	Several promising directions emerge for future work. First, validating our findings on real-world event camera data is essential to assess the sim-to-real transfer capability and practical utility of our dataset. Second, our current work focuses on linear camera motion; exploring rotational and accelerated motions, as well as varying object movement speeds, could enhance the dataset's realism and applicability. Finally, exploring advanced neural architectures --- particularly transformers --- may yield additional performance gains for event-based pose estimation, leveraging their ability to model long-range dependencies in sparse, asynchronous data.

    \section*{Acknowledgements}
    This project was funded in part by the European Union’s Horizon Europe research and innovation programme under grant agreement No. 101120726.

    \bibliographystyle{splncs04}
    \bibliography{bibliography}
\end{document}